\documentclass[conference]{IEEEtran}
\IEEEoverridecommandlockouts
\usepackage{cite}
\usepackage{url}
\usepackage{enumitem}
\usepackage{amsmath,amssymb,amsfonts}
\usepackage{algorithmic}
\usepackage{graphicx}
\graphicspath{ {./images/} }
\usepackage{textcomp}
\usepackage{xcolor}
\usepackage{comment}

\newcommand{\vi}{\lor}
\def\BibTeX{{\rm B\kern-.05em{\sc i\kern-.025em b}\kern-.08em
    T\kern-.1667em\lower.7ex\hbox{E}\kern-.125emX}}

\author{\IEEEauthorblockN{ Ivan Porres\IEEEauthorrefmark{1}, Sepinoud Azimi\IEEEauthorrefmark{1}, Sébastien Lafond\IEEEauthorrefmark{1}, Johan Lilius\IEEEauthorrefmark{1},  Johanna Salokannel\IEEEauthorrefmark{2} and Mirva  Salokorpi\IEEEauthorrefmark{2}  }
\IEEEauthorblockA{\IEEEauthorrefmark{1}Faculty of Science and Engineering \\
Åbo Akademi University\\Turku, Finland \\
\texttt{name.surname@abo.fi}} 
\IEEEauthorblockA{\IEEEauthorrefmark{2}Novia University of Applied Sciences\\ Turku, Finland  \\
\texttt{name.surname@novia.fi}
}
}

\begin{document}

\title{On the Verification and Validation of AI Navigation Algorithms}

\maketitle

\begin{abstract}
This paper explores the state of the art on to methods  
to verify and validate  navigation algorithms for autonomous surface ships. We perform a systematic mapping study to find  research works published in the last 10 years proposing new algorithms for autonomous navigation and collision avoidance and we have extracted what verification and validation approaches have been applied on these algorithms. We observe that most research works use simulations to validate their algorithms. However, these simulations often involve just a few scenarios designed manually. This raises the question if the algorithms have been validated properly. To remedy this, we  propose the use of a systematic scenario-based testing approach to validate navigation algorithms extensively.
\end{abstract}

\section{Introduction}

Maritime Autonomous Surface Ships (MASS) of the future will exhibit an increasing range of self-sufficiency. Autonomous capabilities include relieving the vessel operator from constant supervision by taking over certain responsibilities of the vessels using partial or complete remote operation of vessels, or partial or complete unsupervised navigation. 

An important motivation for autonomous functions and increased intelligence in ships is to improve safety, efficiency of operations and decrease the environmental footprint.  Despite the advances in technologies and constant striving towards improved safety, accidents still happen.   In 2017 alone, 3301 accidents were reported by the European Maritime Safety Agency and over 53\% of all reported accidents were collisions, contacts or grounding occurrences, all due to navigational error
~\cite{EMSA}. The development of autonomous navigational capabilities is seen as a possible solution to dramatically reduce the number of accidents due to navigational error. 

The use of autonomous navigational functions in vessels raises however the question of what may happen if these autonomous functions have design defects. This question is addressed by Valdez et al.~\cite{DBLP:journals/ress/BandaKGGBK19} who present a hazard analysis for the design phase of autonomous vessels. In this study, the authors identify AI software failure as a hazard that can lead to many of the identified accidents.  Valdez proposes a number of safety controls to eliminate or reduce the likelihood that software hazard appears but this study does not address how to implement these safety controls. If we intend to use AI software components in navigation algorithms, we must ensure that they work as expected and we should be able to analyze and reveal whether these components may contain faults. 

Traditionally, navigation algorithms have been based on path planning and optimization  and have been designed manually. Programming is a notoriously complex task and developing defect-free programs require the application of correct by construction methods or an extensive verification and validation effort.

An alternative to path planning and optimization algorithms is the use of machine learning (ML), reinforcement learning (RL) and neural networks (NN). Machine learning has shown staggering success in autonomous cars. Machine learning is known to succeed and outperform traditional approaches specially in vaguely defined problem domains, where it is difficult, if not impossible, to create a full formal specification of the phenomenon under study. We consider this to be the case for COLREGs-based navigation and we conjecture that a ML-based navigation approach can outperform existing search-based and optimization algorithms.  Still, modern AI software may also contain faults introduced during the learning process of a neural network.  As an example, Katz~\cite{DBLP:conf/cav/KatzBDJK17} has analyzed the deep neural network implementation of the next-generation airborne collision avoidance system for unmanned aircraft (ACAS Xu) and found that several logical requirements did not hold for the system as well as some adversarial perturbations that could lead to erroneous collision avoidance actions.

This paper explores the state of the art related to the methods used 
to verify and validate surface ship navigation algorithms.  For this, we have performed a systematic mapping study to find  research works published in the last 10 years proposing new algorithms and we have extracted what verification and validation approaches have been applied on these algorithms. We have observed that most research works use simulations to validate their algorithms. However, these simulations involve just a few scenarios, often designed manually. Therefore, we  propose the use of a systematic scenario-based testing approach to validate navigation algorithms thoroughly.

We proceed as follows. The design of the mapping studied is presented in Section 2, while its main results are presented in Section 3 and 4. Finally, Section 5 describes the proposal for a method for validation of navigation algorithm  using systematic scenario-based testing.

\section{Study Design}
We have adapted and applied the systematic mapping approach described in~\cite{ petersen2015guidelines} to the autonomous maritime domain. In this study, we first defined the appropriate research questions, then conducted the search for the relevant papers. Consequently, we filtered the obtained papers based on our predefined inclusion and exclusion criteria. The result of our study, eventually, ended in producing a systematic mapping.
\subsection{Research questions}
The first step consists in defining the research question. In this study, we define three main research questions. In order to structure the answer to the main questions, we also defined a few sub-questions. Our research questions (RQs) are as follows.

\begin{itemize}
\item[RQ1] What approaches for navigation or traffic avoidance in autonomous ships have been presented in the research literature? 
\begin{enumerate}[label=(\alph*)]
    \item When and where have they been published?
    \item What are the overall approaches?
    \item Do they involve single ship or a swarm of ships?  
\end{enumerate}
  
\item[RQ2] What are the requirements for these approaches as presented in the research literature? 
\begin{enumerate}[label=(\alph*)]
    \item How the safety is defined?
    \item Are the requirements COLREGs compliant?
\end{enumerate}
\item[RQ3] How are these approaches verified and validated in the research literature? 
\end{itemize}

\subsection{Search Strategy}
The primary search is done in the \emph{Web of Knowledge} database, which includes the core \emph{Web of Science} database as well as several regional databases. The core Web of Science database consists of: \emph{Science Citation Index Expanded (1945-present)}, \emph{Social Sciences Citation Index (1956-present)}, \emph{Arts \& Humanities Citation Index (1975-present)}, \emph{Conference Proceedings Citation Index- Science (1990-present)}, \emph{Conference Proceedings Citation Index- Social Science \& Humanities (1990-present)},  and \emph{Emerging Sources Citation Index (2015-present)}. We opted for the papers published between 2010 and 2020. We defined the following criteria for our primary search.

\begin{center}
\textit{
(maritime $\vi$ marine $\vi$  ship $\vi$  vessel) $\wedge$ (autonomous navigation $\vi$  autonomous traffic avoidance $\vi$  collision avoidance) $\wedge$  (algorithm $\vi$   AI $\vi$   artificial intelligence $\vi$  machine learning $\vi$   ML $\vi$   optimization $\vi$  optimisation)
$\wedge$  (validation $\vi$   verification $\vi$   testing $\vi$   simulation $\vi$   quality $\vi$   safe $\vi$   safety)
}
  
 This primary search resulted in the collection of 427 papers.  
\end{center}
\subsection{Inclusion and Exclusion}
At this step, we performed a screening process of the papers, considering only relevant papers based on our inclusion and exclusion criteria. The adopted inclusion criteria are: (1) Only peer-reviewed research papers published in a journal or a conference proceeding; (2) Only papers related to the theme of surface maritime vessel’s in their title or abstract or keywords; (3) Only papers that mentioned machine learning or optimization algorithm in their title or abstract or keywords. The exclusion criteria were: (1) Papers mentioning “maritime vessel” in their abstract but that cannot be considered as describing research on autonomy; (2) Papers are duplicates (3) Papers containing keywords related to our study but discovered as false positives (e.g. review papers, studies on underwater vessels). The full list of papers was equally divided between the authors to apply the inclusion/exclusion criteria and filter the relevant papers. The final list of papers after applying the inclusion/exclusion criteria consisted of 132 papers. 

The identified papers in the screening step were then randomly distributed among four authors for the full reading step. As such, each paper was processed by a second author, to avoid bias. The full list of proceed papers could be found in the Appendix.

\section{Data extraction and classification}
\begin{figure}[tbh]
    \centering
    \includegraphics[scale=0.65]{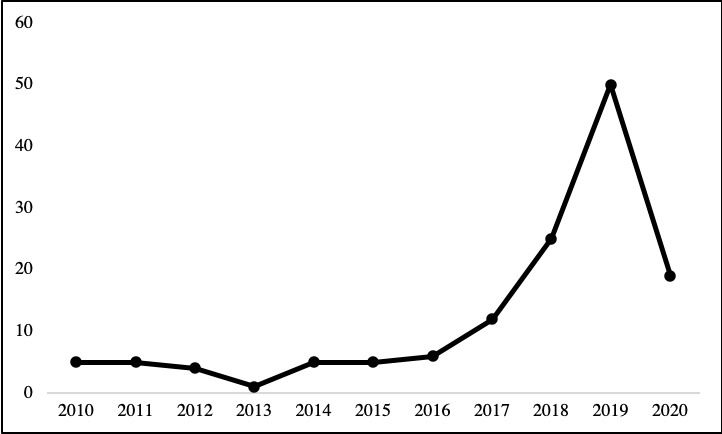}
    \caption{Publication Year}
    \label{fig:year}
\end{figure}

\begin{figure}[tbh]
    \centering
    \includegraphics[scale=0.65]{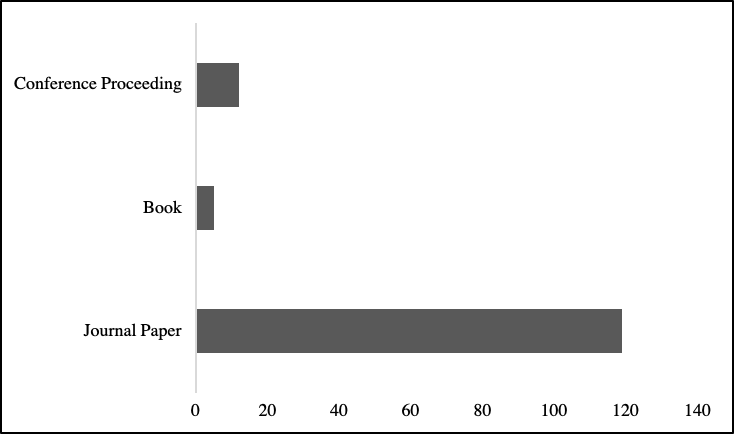}
   \caption{Publication Type}
    \label{fig:type}
\end{figure}

\begin{figure}[tbh]
    \centering
    \includegraphics[scale=0.33]{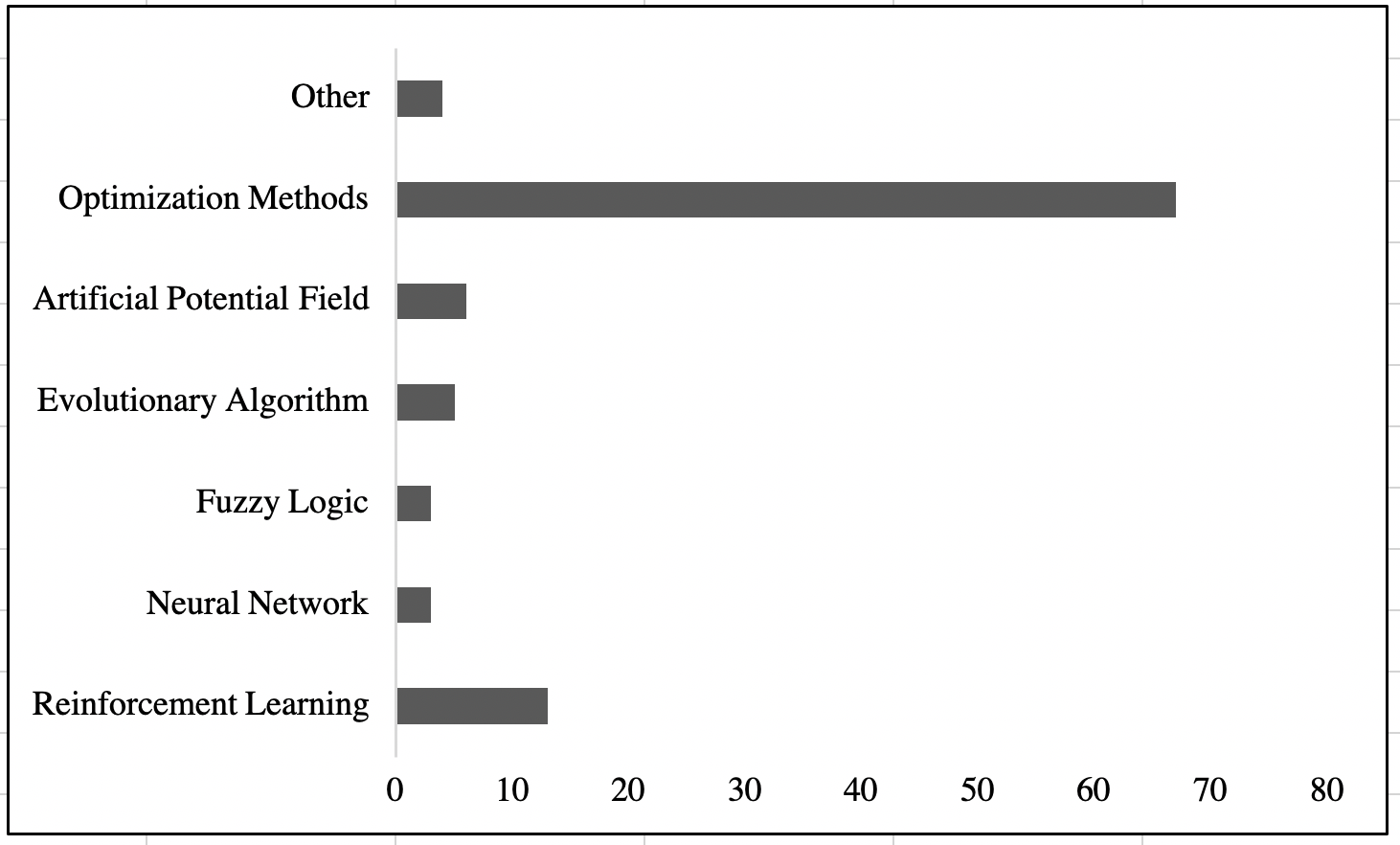}
    \caption{Overall Approach}
    \label{fig:overall}
\end{figure}

For the data extraction we followed the template presented in Table~\ref{tab:data_extraction}. 
\begin{table}[tbh]
\begin{tabular}{|lll|}
\hline
Data Item                         & Value                  & RQ  \\ \hline
\textit{General}                  &                        &     \\
Study ID                          & Integer                &     \\
Paper Title                       & Title of the Paper     &     \\
Authors' Name                     & List of Authors        &     \\
Year of Publication               & Calendar Year          & RQ1 \\
Venue                             & Publication Venue Name & RQ1 \\
\textit{Process}                  &                        &     \\
Overall Approach                  & Algorithmic Approach   & RQ1 \\
Single or Swarm                   & Binary                 & RQ1 \\
Safety                            & Safety Definition      & RQ2 \\
(Non) compliance with Regulations & Binary                 & RQ2 \\
Verification \& Validation        & V\&V Approach           & RQ3    \\\hline
\end{tabular}\caption{Data Extraction Form} \label{tab:data_extraction}
\end{table}
We used the extracted data to answer our main research questions. Figure \ref{fig:year}, \textbf{RQ1(a)}, presents the distribution of the studies between year 2010-2020. As it can be seen from the graphs, the number of publications in the field experienced a dramatic boost in year 2017. The majority of the studies were published as a journal article, followed by conference papers and whole books, 87.5\%, 9\% and 3.5\% respectively, see Figure \ref{fig:type}, \textbf{RQ1(a)}. This is to be expected as the interest in autonomous vehicles have been piqued over the past few years. As it could be observed from Figure \ref{fig:overall}, the majority of the papers opted for optimization as their overall approach, \textbf{RQ1(b)}. This indicates that the use of AI is still at its infancy when it comes to autonomous navigation for maritime surface vessels. Based on the data analysis results, 82\% of the studies involved only one single target ship, whereas the others, focus on a swarm of ships, \textbf{RQ1(c)}.

\begin{figure}[tbh]
    \centering
    \includegraphics[scale=0.65]{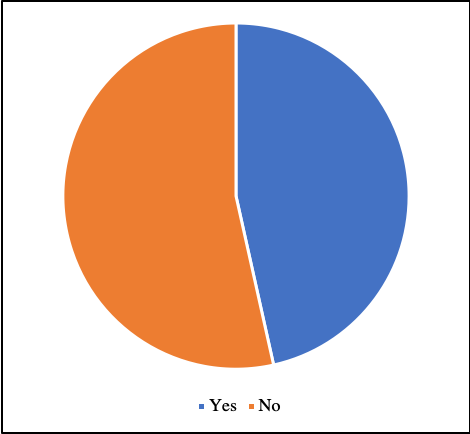}
    \caption{COLREG Compliance}
    \label{fig:colreg}
\end{figure}

The majority of the articles defined safety based on the values of either Time to Closest Point of Approach (TCPA) or Distance to Closest Point of Approach (DCPA), 82\%,  \textbf{RQ2(a)}. Only 48\% of the papers chose to comply with CLOREGs in their study design, \textbf{RQ2(b)}, see  Figure \ref{fig:colreg}.

\begin{figure}[tbh]
    \centering
    \includegraphics[scale=0.65]{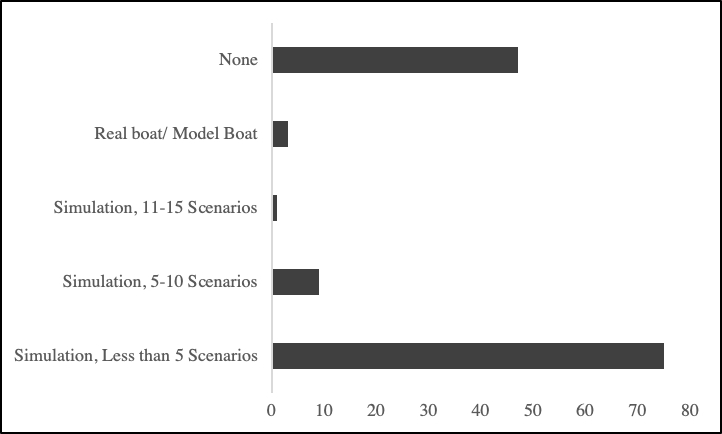}
    \caption{Verification \& Validation Approaches}
    \label{fig:validation}
\end{figure}

The majority of the papers (86 out of 132) identified in this study used simulation approaches to validate their results with a small (ranging from 1 to 12) number of scenarios. Three studies used either a real boat or a model boat for the validation \cite{shen2019automatic,xin2019application,han2020autonomous} and the rest did not use any verification and validation approach, \textbf{RQ3}. The distribution of validation methods is depicted in Figure \ref{fig:validation}

\section{Current practice on the Verification and Validation of AI Navigation Algorithms}

The verification and validation of navigation algorithms is an important issue since software failures has been identified as a hazard that can lead to many accidents in vessels with autonomous functions. To avoid such hazard, Perera proposes a 3-level approach to validate the behaviour of autonomous vessels,~\cite{10.1115/1.4045372}. Level 1 in Perera's classification requires the use of a software simulation for the  motion of all vessels. A level 2 testing system would require that the own ship is a full scale or model vessel that navigates in restricted waters, while the other ships are simulated. In contrast, a level 3 system would require that all involved vessels navigate in open seas.

The mapping study show that most papers use software simulation to validate the proposed results. In these simulations, the simulation starts with a given scenario that describes the initial position and speeds for two or more vessels. The scenario is then animated in the physics-based simulator and the performance of the AI agents under test is evaluated. This corresponds to Level 1 validation in Perera's classification. However, we have observed that most of these works simulate just some few scenarios and that these scenarios are designed manually, often to represent standard situations such as a take over or a crossing. Also, there is a considerable number of research articles that do not contain any verification or validation of their proposed results.

Existing work in the verification and validation in the automotive domain emphasizes the need to use a large number of specially designed scenarios in order to be able to find some faults in autonomous functions. We consider that the same criteria should apply to the maritime domain and that there is a need for domain-specific methods for the  systematic verification and validation of autonomous functions in vessels.  Therefore, we  propose in the next section the use of a systematic scenario-based testing approach to validate navigation algorithms thoroughly.

\section{A Proposal for Navigation Algorithm Validation using Systematic Scenario-Based Testing}

\begin{figure}
\begin{center}
\includegraphics[scale=0.35]{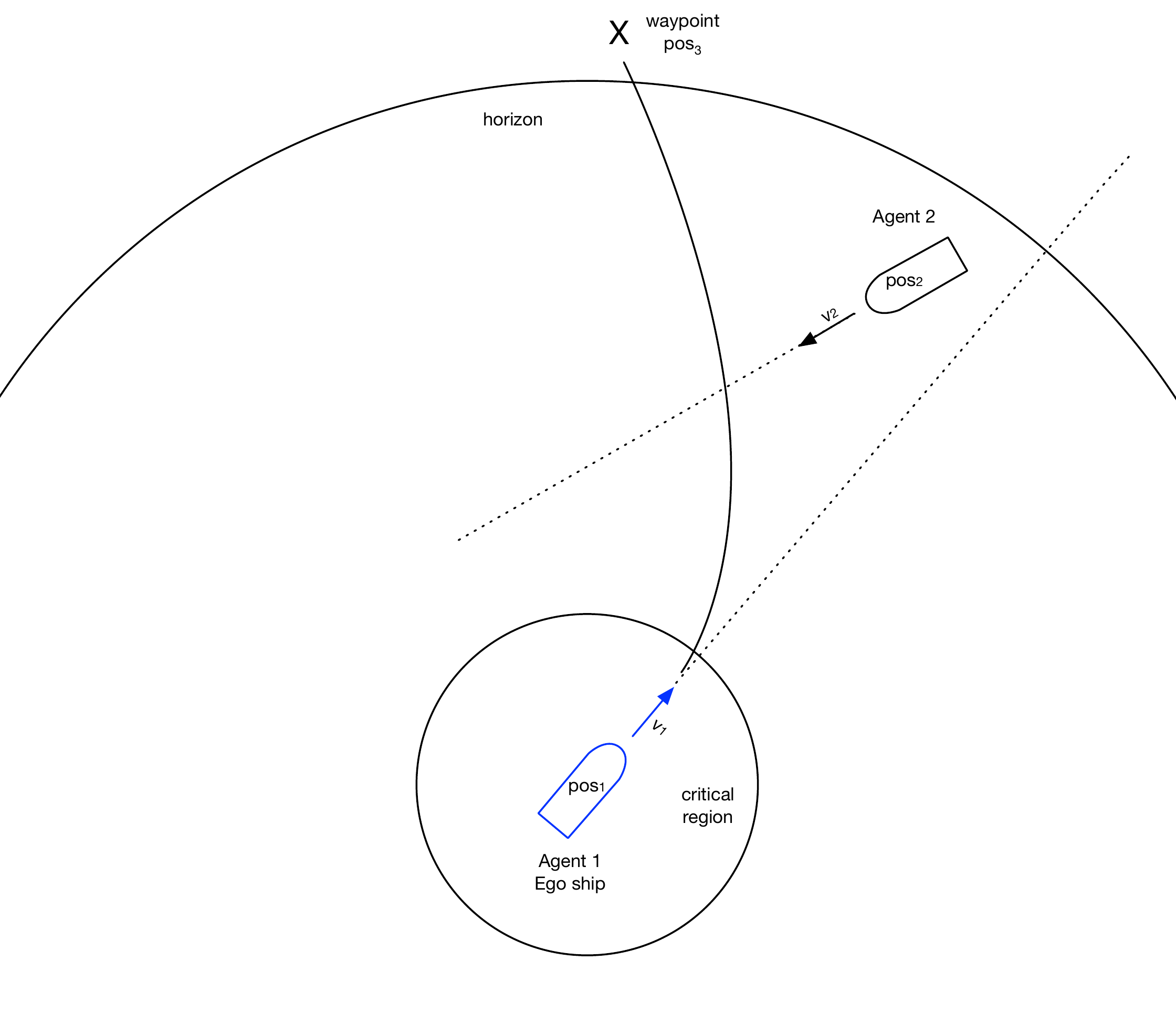}
\caption{A possible scenario}~\label{fig:scenario}
\end{center}
\end{figure}

The goal of scenario-based testing is to evaluate a large set scenarios to find  those where the AI agents do not perform as expected. In each scenario, the position and the velocity vector of each ship may vary, as well as their destination way-point. An example scenario with two vessels is depicted in Fig.~\ref{fig:scenario}.

We are interested to know if there are scenarios where the autonomous validation components under study take the wrong decisions. These are described as \emph{challenging} scenarios in the literature, and they lead to undesirable outcomes such as near miss or a collision.

Testing a single scenario for an autonomous vehicle is computationally expensive since it requires a physics-based simulation in  addition to  executing the autonomous functions. This includes updating the motion of all the vehicles involved in the scenario as well as simulating the  environment sensed by the autonomous functions. Since there is a limited testing budget and we want to maximize the chances to find a defect, it is therefore desirable to select the scenarios that are considered more challenging for the autonomous function,~\cite{DBLP:conf/cdc/GagliardiTR18}.

Several authors have proposed methods to search for challenging scenarios efficiently,~\cite{DBLP:conf/icse/AbdessalemNBS18,DBLP:conf/icra/MullinsSG17}. Abdessalem, Nejati, Vruand and Stifter have proposed a method that uses neural networks as a surrogate model for the scenario fitness functions and then genetic algorithms as a heuristic to search challenging scenarios,~\cite{DBLP:conf/kbse/AbdessalemNBS16}. This is presented as a two phase process. First a set of simulations must be executed in order to create the surrogate models of the fitness functions. Once these models have been created, the scenario search is performed.

We have proposed a new approach for scenario-based testing that it is specific to maritime surface vehicles and that avoid the need of training subrrogate models. Our approach, presented in~\cite{SEAA2020}, is based on the use of a neural network to discriminate and select scenarios that may be challenging for the autonomous system being tested. The selected scenarios are simulated and evaluated and their outcome is used to train the discriminating neural network. Compared to other works such as~\cite{DBLP:conf/kbse/AbdessalemNBS16}, we combine the training of the discriminator network and the scenario selection in one step, with the intention to reduce the number of necessary simulations.  The simulations are evaluated by risk of collision and compliance to COLREGs. 

To evaluate our approach, we have tested a collision avoidance algorithm based on a neural network trained using reinforcement learning. The evaluation task was to create 6000  simulation scenarios, each one depicting a different initial situation. Our experimental results show that the proposed testing method generates test suits composed mostly of challenging scenarios. This allows us to validate quickly if the navigation algorithm under test can operate safely while abiding the COLREGs.

\section{Conclusions}
This paper explores the state of the art on the methods  
to verify and validate  navigation algorithms for autonomous surface ships by carrying out a systematic mapping study. The mapping study reveals that most research works use simulations to validate their algorithms. Finally, we have proposed the use of a systematic scenario-based testing approach to validate navigation algorithms extensively.

\bibliographystyle{ieeetr}
\bibliography{bibliography.bib,all-new.bib}

\appendix\label{sec:List}
\section*{Articles included in the Mapping Study}

\end{document}